\documentclass[acmsmall]{acmart}

\AtBeginDocument{%
  }

\setcopyright{acmlicensed}
\copyrightyear{2024}
\acmYear{2024}
\acmDOI{XXXXXXX.XXXXXXX}

\begin{document}

\title{A Comprehensive Survey of AI-Driven Advancements and Techniques in Automated Program Repair and Code Generation}

\author{Avinash Anand}
\affiliation{%
  \institution{Indraprastha Institute of Information Technology}
  \city{New Delhi}
  \state{Delhi}
  \country{India}
}

\author{Nishchay Yadav}
\affiliation{%
  \institution{Indraprastha Institute of Information Technology}
  \city{New Delhi}
  \state{Delhi}
  \country{India}
}

\author{Akshit Gupta}
\affiliation{%
  \institution{Indraprastha Institute of Information Technology}
  \city{New Delhi}
  \state{Delhi}
  \country{India}
}

\author{Shaurya Bajaj}
\affiliation{%
  \institution{Indraprastha Institute of Information Technology}
  \city{New Delhi}
  \state{Delhi}
  \country{India}
}

\renewcommand{\shortauthors}{Avinash Anand, Nishchay Yadav, Akshit Gupta, Shaurya Bajaj}

\begin{abstract}
Bug fixing and code generation have been core research topics in software development for many years. The recent explosive growth in Large Language Models has completely transformed these spaces, putting in reach incredibly powerful tools for both. In this survey, 27 recent papers have been reviewed and split into two groups: one dedicated to Automated Program Repair (APR) and LLM integration and the other to code generation using LLMs. The first group consists of new methods for bug detection and repair, which include locating semantic errors, security vulnerabilities, and runtime failure bugs. The place of LLMs in reducing manual debugging efforts is emphasized in this work by APR toward context-aware fixes, with innovations that boost accuracy and efficiency in automatic debugging. The second group dwells on code generation, providing an overview of both general-purpose LLMs fine-tuned for programming and task-specific models. It also presents methods to improve code generation, such as identifier-aware training, fine-tuning at the instruction level, and incorporating semantic code structures. This survey work contrasts the methodologies in APR and code generation to identify trends such as using LLMs, feedback loops to enable iterative code improvement and open-source models. It also discusses the challenges of achieving functional correctness and security and outlines future directions for research in LLM-based software development.\newline
\end{abstract}

\ccsdesc[500]{Computing methodologies~Artificial intelligence}
\ccsdesc[500]{Software and its engineering~Software creation and management}

\keywords{Code Generation, Automated Program Repair}


\maketitle

\section{Introduction}
Large Language Models (LLMs) have steadily gained popularity in the field of automated software engineering, including primarily bug fixing \cite{timing_side_channel2024} \cite{greybox_fuzzing2024} \cite{provenfix2024} \cite{fuzzing2024} \cite{protocol_fuzzing2024} and code generation \cite{openai2021} \cite{codet5_2021} \cite{graphcodebert2021}. Over the past decade, APR and code generation have greatly risen in use \cite{multi_agent2024} \cite{automatic_programming2024} and hence, have also warranted a great amount of research into the subject. A lot of tools that employ APR and natural language processing for code generation have been developed \cite{openai2021} \cite{codet5_2021} \cite{graphcodebert2021}, which use an array of different techniques, including implementing Abstract Syntax Trees (ASTs), using varying heuristics for ranking plausible patches, patterns, and context-matching, among others. Using LLMs in code-related tasks has significantly improved the quality and speed of automating programming and discovering bugs in code. These tasks include summarizing code, generating code based on natural language requests, fixing bugs in pre-existing code, and understanding relatively large and complex repositories. However, in this paper, we’ll only be covering the research and studies done in the field of code generation and bug fixing, and to make understanding the work done in these fields easier, we’ve divided the tools and papers that we’ve covered into these two categories. LLMs have widely gained usage in these tools due to the natural advantage of having been trained on extremely large datasets and billions of parameters. Hence, it is easier to employ large LLMs to do particular tasks pertaining to programming. This leads to impressive performance and sizable advantages when compared to training models from scratch \cite{timing_side_channel2024} \cite{provenfix2024} \cite{fuzzing2024}. Simultaneously, the task of employing LLMs in APR and code generation is also extremely complex and encompasses various areas with their own extensive research, such as benchmarking, repair scenarios (syntactic errors, semantic errors, etc.), repair techniques (recompilation, binary rewriting, etc.) testing of repairs (patch generation, input testing, coevolution) among others. Hence, understanding the work already done in this field can prove to be quite complex and time consuming. 
This survey paper aims to summarize the research and work already accomplished in this growing field to assist others in gaining a better understanding of how these tools work, their performance in practical scenarios, the areas they work on, and their limitations. We’ve collected 27 papers and summarized various factors about them, including the LLMs they use, the programming languages they work on, and, by extension, the difficulties faced in building language-agnostic APR tools, the approach they take to repair bugs and generate code, and the challenges that are still being worked on in the field.
In conclusion, this paper aims to:

\begin{enumerate}
    \item Collect research done on APR and code generation using LLMs and summarize the goals achieved.
    \item Elucidate the repair scenarios these tools can be used for and the programming languages they work on.
    \item How LLMs are integrated into the workflow of repairing and generating code and the challenges faced in doing so.
    \item The limitations of using LLMs in code-related tasks and cases which are still being worked on.
\end{enumerate}

\section{Survey Methodology}
This section covers the measures taken in order to carry out the survey, the ways which we have implemented in to search, collect and filter out the models, research papers and journals which closely align with our purpose for this literature survey. 

\begin{figure}[htp]
    \centering
    \includegraphics[width=10cm]{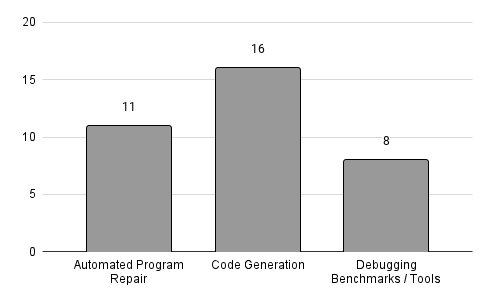}
    \caption{Papers Included in Every Domain}
    \label{fig:graph}
\end{figure}

\subsection{Research Questions}

Through this survey we aim to answer the following research questions, our criteria and metrics for inclusion of a particular paper focuses mostly on whether it has helped us to answer any of these questions more comprehensively. Any resource that added no value on any of these topics were discarded from the study. All of these questions revolve around the existing and the emerging technologies around APR and Code generation
\begin{enumerate}
    \item How have AI techniques, especially large language models (LLMs), improved software debugging and bug fixing? What are some recent trends and common challenges in using AI for these tasks?
    \item How do modern debugging tools and benchmarks help evaluate the effectiveness of bug-fixing methods? What are the common gaps or limitations in these tools?
    \item What are the key differences between popular code generation models? How do these models perform on tasks like code completion and code summarization?
    \item How do different AI models and tools compare in handling various code-related tasks? What are the main strengths and weaknesses of these models?
\end{enumerate}

\subsection{Methods Employed}
A number of methods were used in order to select and bifurcate the resources so that the survey can be done more effectively, these are the following : 
\begin{enumerate}
    \item \textbf{Systematic Literature Review} \\
    A review of the available literature on the subject of interest and related work. This also involves having a clear inclusion and exclusion criteria. It is done using searching relevant papers revolving around the main topic among various databases. Then applying inclusion criteria and exclusion criteria mentioned above and finally conducting a detailed analysis and categorizing the selected papers/models under umbrella categories of Code Generation and  Automated Program Repair as shown in Fig. 1.
    
    \item \textbf{Taxonomy Development} \\
    This method is employed by creating a classification of the AI techniques, tools and methods used for debugging, bug fixing and code generation. This comparative analysis is done upon the objectives, techniques and outcomes, further figuring out overlaps and distinctions between categories.

    \item \textbf{Comparative Analysis} \\
    A comparison was made between the selected papers using various evaluation criteria such as performance, accuracy etc. Tabular and graphical description for a better visualization of the comparison was done between the model and tools.

    \item \textbf{Trend Analysis and Gap Identification} \\
    A dedicated survey objective was to identify open challenges, and research gaps in AI-driven techniques and methods for APR and code generation. In a category specific approach it was attempted to figure out the common themes, recurring challenges, and under-explored areas across all categories. An attempt to see the future trends and potential in the similar field was made as well.

    \item \textbf{Survey of Benchmarks and Evaluation Metrics} \\
    A part of the survey involves a study on the benchmarks and evaluation metrics employed for evaluating the models and tools. Identifying similarities and differences in the benchmarks and listing out the description for each of them ultimately analyzing the strengths and weaknesses of different benchmarks and suggesting improvements or new metrics.
    
\end{enumerate}

\section{Research Questions}

\subsection{How have AI techniques, especially large language models (LLMs), improved software debugging and bug fixing? What are some recent trends and common challenges in using AI for these tasks?}

Automated Program Repair (APR) can be used to achieve a number of goals, one being how it can be applied to fix security bugs through numerous approaches. Security bugs often arise from vulnerabilities like buffer overflows, input validation issues, race conditions, or improper access control. APR can be used to address these issues in different ways, depending on the type of bug, the repair technique, and the tools used. Below are several ideas on how APR can be applied for fixing security bugs:

\subsubsection{\textbf{{Security Bugs}}}
\begin{enumerate}
    \item
    \textbf{Template Based Patching:} There are many Automatic Program Repair (APRs) tools that use already existing templates to fix security vulnerabilities \cite{timing_side_channel2024} \cite{provenfix2024}. For instance, they can fix SQL injection and other types of bugs. Thus, the process is made faster when at least some bugs are already fixed.
    
    \item
    \textbf{Dynamic Analysis for Security Bugs:} Methods like fuzzing and symbolic execution are the reasons for security flaws that might have been unnoticed at the time of the execution of the program \cite{protocol_fuzzing2024}. Afterward, the tools can generate dummy patches which will be tested by executing on various inputs. The results of the testing will then enable the patches and input to co-evolve, thus, the fix for the problem will be improved \cite{evolutionary_testing} \cite{fuzzing2024} \cite{protocol_fuzzing2024}.

    \item
    \textbf{Search-Based APR for Security:} Search-based methods are widely utilized in different APR and debugging systems to create and confirm patches based on mutations of the initial code \cite{fuzzing2024}. These tools likewise act in the same way as the previous ones, where they output a list of likely patches and then use different methods to search for the optimal patch that meets the given conditions \cite{evolutionary_testing} \cite{fuzzing2024}.

    \item
    \textbf{Specification-Guided APR for Security Protocols:} The APR tools we looked at mostly used fuzzing to verify the security of different internet protocols. This process was done by obtaining precise grammar and specifications for the protocols \cite{fuzzing2024}. Then the patches are checked by applying them on different kinds of inputs and they are tested on the protocols \cite{evolutionary_testing} \cite{provenfix2024} \cite{fuzzing2024}.

    \item 
    \textbf{Input Sanitization and Validation Patches:} The problem of security vulnerabilities in software may be due to the fact that there is no proper input validation or sanitization. Such a situation can lead to problems such as injection attacks or data corruption. Automated Program Repair (APR) tools can detect these flaws by checking how inputs are treated and processed in the code. One approach is to detect the lack of sufficient input validation after which, the tool can automatically add the required functions for sanitizing, filtering, or escaping harmful characters, for example. Through this method, APR enhances the security of the software, preventing the use of harmful inputs to breach the system or cause the software to behave unexpectedly. Besides, this automated method reduces the manual intervention, thus, ensuring more consistency and reliability of input validation throughout the codebase \cite{fuzzing2024}.

    \item 
    \textbf{Repairing Memory Safety Bugs:} Bugs concerning memory safety, such as use-after-free, buffer overflows, and null pointer dereferences, are very common in low-level programming languages, particularly C and C++. These security threats can bring about situations of high risk, crash, and non-deterministic behavior in the software. Automated Program Repair (APR) methods can both detect unsafe memory accesses \cite{provenfix2024} and apply corrective measures, such as conducting bounds checking, utilizing safer memory allocation techniques \cite{protocol_fuzzing2024}, or replacing raw pointers with smart pointers. The positives of automatically identifying and treating these issues are, APR helping to improve the robustness and security of programs, and thus, the exploitation or failures probability decreasing.
    
\end{enumerate}

\begin{figure}[htp]
    \centering
    \includegraphics[width=10cm]{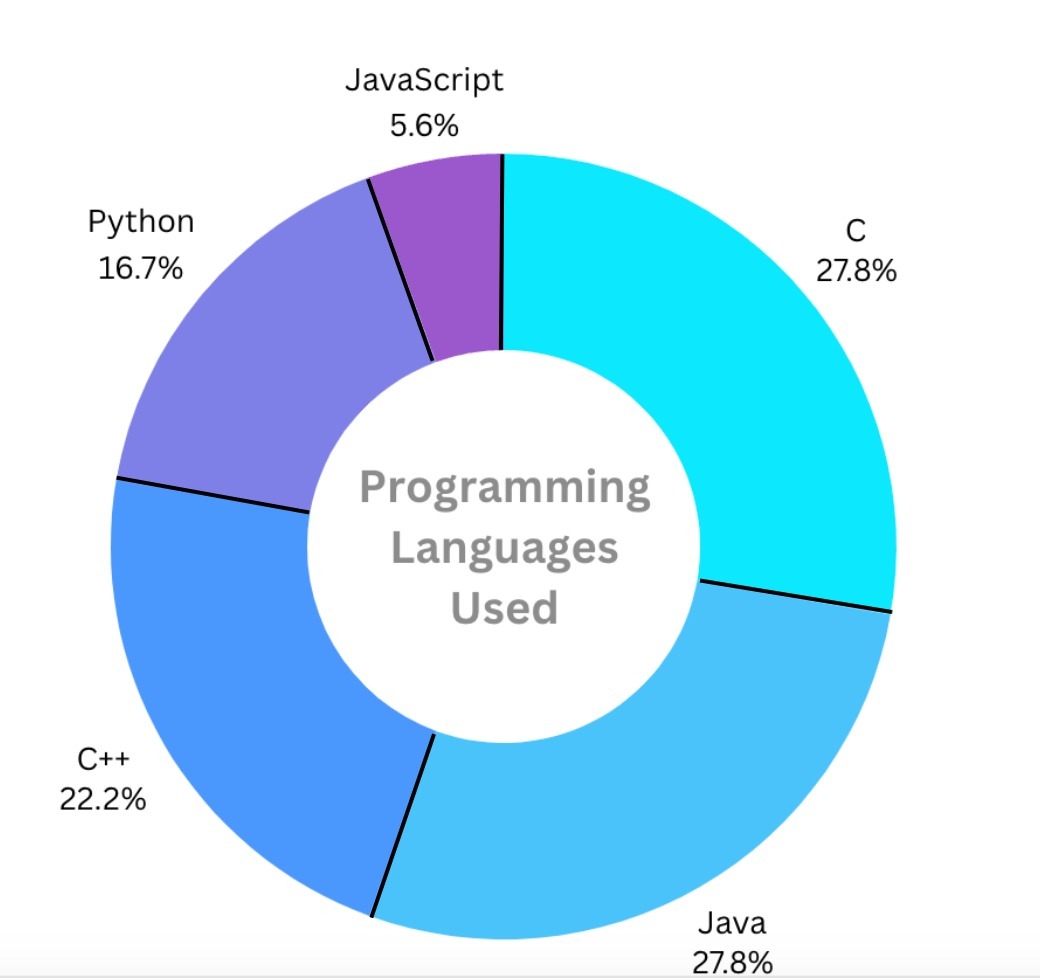}
    \caption{Programming languages in the papers we surveyed}
    \label{fig:graph}
\end{figure}

\subsubsection{\textbf{{Semantic Bugs}}}
\begin{enumerate}
    \item
    \textbf{Pattern-Based Patching for Semantic Bugs:} The repair of semantic bugs, which consist of logical errors or deviations from the expected functioning of the program, can be addressed through pattern-based Automated Program Repair (APR) systems \cite{opencodeinterpreter2024} \cite{provenfix2024} \cite{protocol_fuzzing2024}.
    
    \item
    \textbf{Fuzzing Techniques for Semantic Bugs:} The fact that fuzzing is a very useful tool to discover memory and semantic bugs is a development in psychometrics. Techniques like Greybox Fuzzing \cite{greybox_fuzzing2024} which is a testing technique where the fuzzer has partial knowledge of the program's internal structure, combining both black-box and white-box testing elements, are found among such techniques. Its goal is to generate test inputs that are as efficient as possible in revealing the bugs using the feedback from the program's execution. This fuzzing method enables the detection of bugs that are very difficult to uncover by the traditional testing methods \cite{fuzzing2024} \cite{protocol_fuzzing2024}.

    \item
    \textbf{Search-Based APR for Semantic Bugs:} Using evolutionary algorithms, it is possible to search different repair options with search-based APR techniques \cite{provenfix2024} \cite{fuzzing2024}. In this case, the candidates produced are also the ones that are covered by the test suites or specifications to be able to find the fixes that remove semantic inconsistencies. The process includes mutating a part of the program, running the variants, and picking the most efficient one. Search-based APR has been very successful in correcting semantic errors for which a specific bug pattern is not easy to detect.

    \item
    \textbf{Specification-Guided Repair:} Semantic bugs frequently happen due to incorrect or incomplete specifications \cite{deepseek2024} \cite{provenfix2024}. APR systems that use specification-guided repair rely on formal models of the expected behavior to identify the deviations \cite{fuzzing2024}. This method is practiced today because of the fact that the patches produced not only solve the bugs but also conform to the formal specifications of the system \cite{protocol_fuzzing2024}.
    
\end{enumerate}

\subsubsection{\textbf{{Syntactic Bugs}}}
\begin{enumerate}
    \item
    \textbf{Pattern-Based Patching for Syntactic Bugs:} Syntactic bugs, which are the bugs that pertain to programming language syntax rules as well as having other issues related to incorrect code structures, can be dealt with through pattern-based Automated Program Repair (APR) systems in a very efficient manner. These systems consider the syntactic violations and then use the predefined correction patterns to get to the desired outcome. FixMiner is a tool that uses mining previous bug-fixing commits of common syntax errors to generate repair patterns, based on those corrections, offering an effective solution for recurring syntactic issues \cite{tabbyml2024}.
    
    \item
    \textbf{Grammar-Based Fuzzing for Syntax Validation:} Tools that use grammar-based fuzzing are concerned with syntactic bugs and they therefore produce program inputs which adhere to the grammar rules of the language in question. These tools modify the inputs according to the Abstract Syntax Trees (ASTs) to keep them syntactically valid, which in turn helps to get bugs that have to do with incomplete or wrongly structured code. Grammar-aware mutation strategies have proven effective in discovering syntactic anomalies that can disrupt program execution \cite{deepseek2024}.

    \item
    \textbf{Search-Based Techniques for Syntactic Repairs:} Search-based techniques can be applied to fix syntactic bugs in a similar manner as their use for semantic ones. The methods in this case look through many code modifications and then test them so as to get a valid syntactic fix. These tools explore a space of potential syntactic mutations and create candidate patches that are consistent with the programming language's syntactic rules and the program's original intent.
    
\end{enumerate}

\subsection{What recent trends have emerged for using AI in APR tools?}
Recent Trends:
\begin{itemize}
\item Pre-trained Models: The recruiting of the use of pre-trained models (Codex, CodeT5) that have been fine-tuned on large programming datasets is gaining traction.

\item Transfer Learning for Code: LLMs are becoming more adaptable to transfer learning, which is a technique where models pre-trained on general code can be fine-tuned for specific bug-fixing tasks.

\item Self-Supervised Learning: This allows LLMs to train on unlabeled datasets, taking advantage of the fact that there are a lot of code reposito ries without needing someone to annotate them.

\item Explainable AI (XAI): There are initiatives to make the AI-powered bug fixing process more interpretable and transparent so that developers can understand the reasoning behind the AI's solution.

\item Interactive Debugging Systems: Tools are becoming more interactive, integrating active learning where AI requests human help to resolve unclear cases.

\item Multi-modal Models: New models are not only incorporating code but also comments, documentation, and logs, which enhances the AI's ability to understand the context of bugs.
\end{itemize}

There have been a large number of advancements in the realm of automated program repair which have streamlined the process of integrating artificial intelligence and large language models into code-related tasks. These include:

\begin{enumerate}
    \item
    \textbf{Pre-trained models:} The use of pre-trained models such as GPT-4 which have been fine-tuned on large programming and bug datasets \cite{protocol_fuzzing2024} has shown a steady upward trend over the past few years.
    
    \item
    \textbf{ Neural Networks for Fault Localization:} Through artificial intelligence, a program can carry out a deep analysis of some properties of the code like temporal properties \cite{provenfix2024} and thus be able to identify the bugs in the program. A number of APR tools also make use of Abstract Syntax Trees (ASTs) \cite{timing_side_channel2024} \cite{evolutionary_testing} \cite{fuzzing2024} to be able to reason about the control flow of the program and fixing it.

    \item
    \textbf{Automated Test Generation:} Using AI to automatically create and run test cases \cite{evolutionary_testing} \cite{fuzzing2024} in order to check the correctness of the potential solutions, so that the repairs are not breaking the existing functionality and also meet the desired specifications.

    \item
    \textbf{Test Coverage Improvement:} Incorporating AI to dissect code modifications and guarantee that new or revised tests encompass all pertinent elements of the codebase, thus boosting the accuracy of the automated repairs \cite{evolutionary_testing} \cite{fuzzing2024} . This also brings down the chances of overfitting which can be a cause of unexpected results for a program \cite{fuzzing2024}.
\end{enumerate}

\begin{figure}[htp]
    \centering
    \includegraphics[width=10cm]{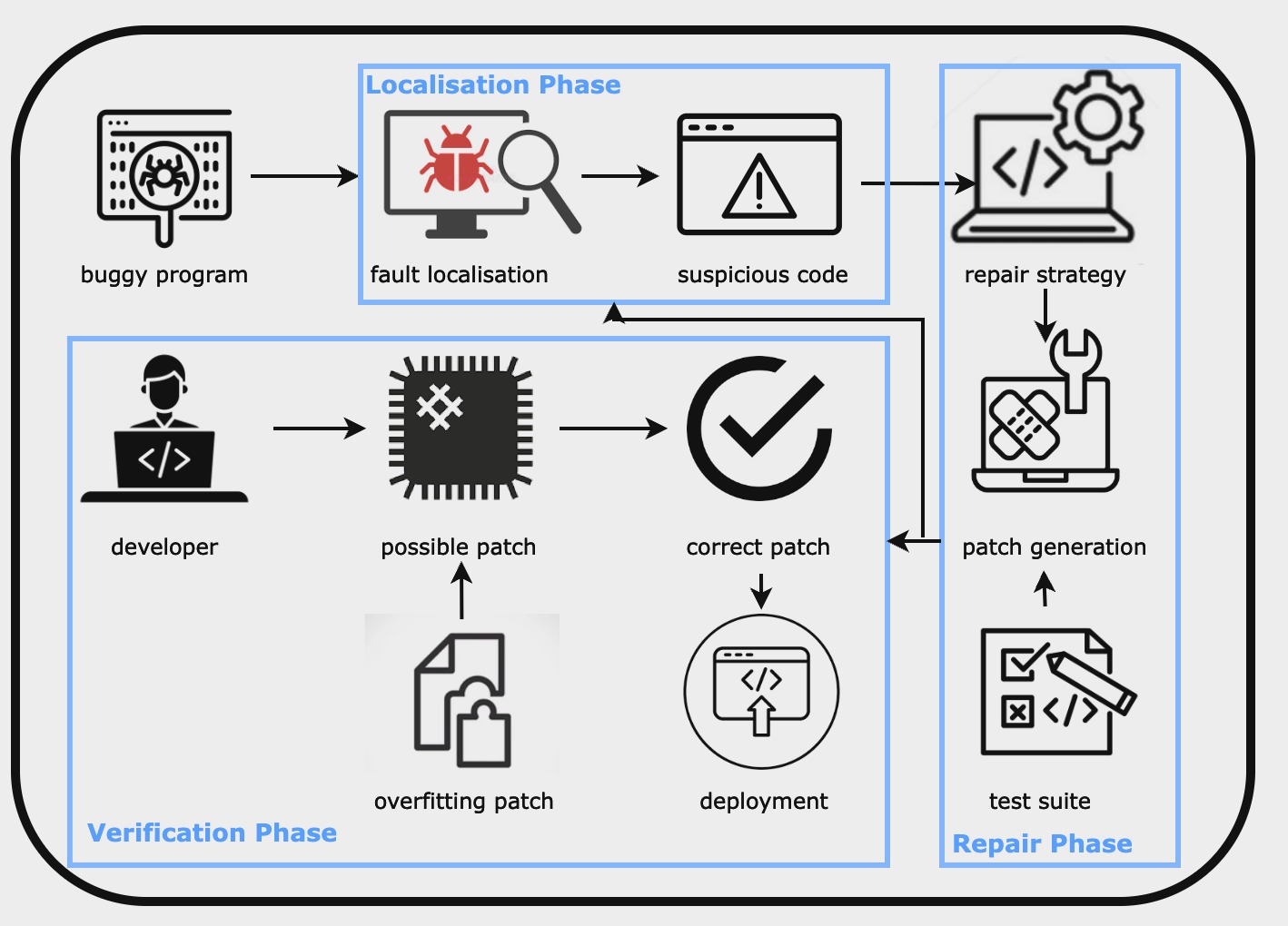}
    \caption{The workflow of automated program repair tools}
    \label{fig:graph}
\end{figure}

\subsection{What challenges are being faced in APR right now?}
In spite of having at least hundreds of studies on the subject and AI tools developed to the highest level, APR is still having some problems and challenges that are yet to be solved. These include:
\begin{enumerate}
\item \textbf{Accuracy and Reliability:} APR tools still face the problem of sometimes incorrectly identifying perfectly fine code as faulty and vice-versa \cite{greybox_fuzzing2024}. Hence, the corrections made by APR tools often need to be verified by humans before being implemented.

\item \textbf{Context Sensitivity:} APR tools still have difficulties with the comprehension of large codebases that have a lot of dependencies. This is one of the main reasons for the use of incorrect practices and repairs which are correct from the technical point of view but not for the whole codebase \cite{timing_side_channel2024}.

\item \textbf{Resource Overhead:} APR tools also have high demands regarding memory and computing power to work adequately \cite{software_repository2024}. Hence, the resource overhead may sometimes be the reason for the user’s workflow disruption and thus the productivity decrease.
\end{enumerate}

Common Challenges:

\begin{itemize}
    \item
    Generalization: LLMs frequently fail to generalize to new, previously unseen, bugs or highly domain-specific code, especially in systems with unconventional architectures or libraries.
    
    \item
    Scalability: Debugging large, complex systems is still one of the most difficult tasks for LLMs as they have to deal with a huge amount of potential interactions and dependencies in the code.

    \item
    Limited Understanding of Context: AI models have gone through many improvements but they may still make mistakes in understanding the full business logic or domain-specific intricacies behind the bugs which may result in incomplete or incorrect fixes.

    \item
    Security Concerns: AI-generated fixes may inadvertently introduce security vulnerabilities or fail to address existing ones.

    \item
    Bias in Training Data: LLMs may inherit biases from the data they were trained on, leading to over-reliance on common patterns and overlooking edge cases.

    \item
    Overfitting to Benchmarks: Models that are trained mostly on benchmarks might specialize in certain datasets and hence not perform well in real-life scenarios.

    \item
    Ethical Concerns: Copyright issues arise when LLMs are trained on proprietary code, and there are concerns about reproducing code without proper credit.
\end{itemize}

 \renewcommand{\arraystretch}{1.75} 

 \begin{table}[]
 \caption{Targeted programming languages in the surveyed papers}
 \resizebox{\textwidth}{!}{
 \begin{tabular}{|l|l|l|}
 \hline
 \multicolumn{1}{|c|}{\textbf{Paper Title}}                            & \multicolumn{1}{c|}{\textbf{Programming Languages Targeted}} & \multicolumn{1}{c|}{\textbf{Benchmark(s)}} \\ \hline
 Report on Timing Side-Channel Mitigation via Automated Program Repair & C                                                            & QFuzz benchmark suite                      \\ \hline
 Report on Greybox Fuzzing for Concurrency Testing                     & C, C++                                                       & SCTBench, ConVul                           \\ \hline
 Coevolution of Patches and Tests in Automated Program Repair          & Java                                                         & Defects4J                                   \\ \hline
 ProveNFix: Temporal-Guided Program Repair                             & OCaml                                                
& Custom Benchmark                             \\ \hline
 Program Repair by Fuzzing over Patch and Input Space                  & C                                                            & VulnLoc                                    \\ \hline
 LLM-guided Protocol Fuzzing                                           & C                                                            & ProFuzzBench                               \\ \hline
 DEAR                                                                  & Java                                                         & Defects4J                                  \\ \hline
 Transfer                                                              & Java                                                         & Defects4J                                  \\ \hline
 SPT Code                                                              & Java, Python, JavaScript, PHP, Go, Ruby                          & BLEU, METEOR, ROUGE-L                             \\ \hline
 Debug like a Human: A Large Language Model Debugger                   & Python, Java, C, C++                                         & HumanEval, MBPP, TransCoder                \\ \hline
 Evaluating Debugging Capability of Large Language Models              & Python, C++, JavaScript, Java                                & DebugBench, HumanEval, MBPP                \\ \hline
 \end{tabular}
 }
 \end{table}

\section{How do modern debugging tools and benchmarks help evaluate the effectiveness of bug-fixing methods? What are the common gaps or limitations in these tools?}
Benchmarking is an important part of exploring new possibilities for APR tools and identifying the repair scenarios that can be used. Several of the papers we surveyed performed a rigorous exercise in benchmarking and comparison with existing state-of-the-art debugging tools. Benchmarking against this baseline can help researchers and developers explore new ways to improve their solutions and hopefully expand the set of scenarios we can trust APR to successfully fix.
\subsection{Modern Debugging tools}
In software engineering, modern tools of debugging assist the validation of bug-fixing approaches mainly by automating the defect detection, location, and repairs on the code. Some of them, including FuzzRepair \cite{fuzzing2024} and AFLNet \cite{protocol_fuzzing2024}, integrate code fuzzing techniques, coupled with application of program repair methods, to stage from one part of the code to another, generating diverse yet reachable inputs leading them to bugs located in the hard-to-reach areas. Such approaches based on fuzzing are frequently employed to assess how resilient bug-fixing methods are, because all of them directly run patched code under diverse conditions to check how thoroughly the fix holds in a variety of scenarios-and provide information that is valuable in supporting the robustness of the repairs.

With regard to machine learning, deep learning techniques for automatically generating patches have been developed based on existing bug-fixing patterns such as CoCoNuT, SequenceR, and Tufano19 \cite{dear2024}, among others. These tools make well-supported research in assessing how well a bug-fixing method generalizes across different codebases. For instance, CodeBERT, GraphCodeBERT, and CugLM \cite{spt_code2024} evaluate bug fixes by embedding richer semantic code representations relative to each other to allow for better targeting of potential faults and testing the effectiveness of repairs through natural language processing and code-understanding models. TreeBERT and T5 \cite{spt_code2024} further expand on this development by learning and understanding the structural relationships within code as well as ensuring that bug fixes are not only syntactically correct but also logically consistent with the rest of the system.

ARJA-e, REWARDREPAIR, and EVOREPAIR \cite{evolutionary_testing} are tools for evolutionary and template-based repair that use genetic algorithms and reward-based learning to automatically generate and validate patches. These tools are quite important for evaluating bug-fixing methods as they can measure the efficiency and scalability of repair techniques, especially in complex codebases. Similarly, TBAR and Darjeeling \cite{fuzzing2024} generate fixes based on predefined templates, which can then be utilized to analyze the effectiveness of rule-based or human-guided patch generation. Prophet, on the other hand, uses probabilistic models to predict the correctness of the patches. Thus, it proposes a method to evaluate the success probability of a bug fix before applying it in a production environment \cite{fuzzing2024}.

Tools like Fix2Fit and CPR \cite{fuzzing2024} (Conditional Program Repair) make sure that not only the rightness of fixes is checked but also their "fit" in the whole program context, which in turn decreases side effects that might add up to new bugs, are the other example of the explained practice. SAVER and DLFix \cite{dear2024}, with their emphasis on semantic analysis and deep learning models, make sure that bug fixes not only fix the immediate problem but also secure the integrity of the program as a whole. The use of these tools by the developers is often to ensure that the bug fixing methods are put under tough tests for robustness, efficiency, and scalability, thus, improving software reliability and stability.

In conclusion, these tools give a range of methods like fuzzing, template-based repair, machine learning, and semantic analysis, which together can provide a thorough assessment of bug-fixing techniques as well as checking for its accuracy, robustness, and applicability.

\subsection{Modern Benchmark tools}
Using modern-day benchmarks is invaluable for measuring the efficiency of bug-fixing techniques that require standardized and controlled settings in which researchers will be able to run the testing and verification procedures as thoroughly as possible. These benchmarks include different functions like checking code generation abilities or measuring debugging efficiency among the vulnerability detection and debugging.
In a precise manner, bug-fixing performance tests should be conducted within the standards of software quality, which are the different dimensions of software quality. Code generation benchmarks like HumanEval \cite{debug_like_human2024} \cite{evaluating_debugging2024} and MBPP \cite{debug_like_human2024} \cite{evaluating_debugging2024} are fundamental for this purpose. HumanEval challenges models with a set of programming problems to determine how well they can generate correct code and fix issues. At the same time, MBPP offers an array of programming solutions to the problems, which are then used to check the efficiency of the bug-fixing methods in generating and correcting code. These benchmarks help to the decision of technical meaning to what degree bug-fixing techniques can work together with code generation processes and fix real-world problems.
For example, Fuzzing benchmarks such as ProFuzzBench \cite{protocol_fuzzing2024} and SCTBench \cite{greybox_fuzzing2024} test the bug-fixing methods in terms of their efficiency in vulnerability detection and addressing of vulnerabilities. ProFuzzBench's network protocol implementations and fuzzing tools lend a hand in debugging experiments and help researchers find out if bug-fixing methods are resistant to the vulnerabilities of network protocols. SCTBench uses its collection of multithreaded benchmarks to detect concurrency problems, thus verifying how well bug-fixing techniques can deal with synchronization and multithreading problems.
Debugging benchmarks like DebugBench \cite{evaluating_debugging2024} and VulnLoc \cite{fuzzing2024} are testing different debugging methods and the algorithms that can be used. DebugBench assesses large language model debugging task performances, revealing important information about the use of these models when trying to identify and correct errors. In contrast, the process of VulnLoc has the focus on automatic vulnerability localization which implies finding out the root cause of bugs with the smallest possible error margin.
Moreover, Defects4J \cite{evolutionary_testing} \cite{dear2024} \cite{fault_localization2024} supplies defect data composed of real Java bugs through a database that is full of Java bugs. This benchmark supports the testing of bug-fixing methods against real defects, therefore, bringing the necessary practical perspective on the effectiveness of the solutions. TransCoder \cite{debug_like_human2024} is also somehow important because of its code translation between programming languages, so that researchers can do the correctness and bug test during the code translation with the bug-fixing techniques.
To sum up, modern benchmarks are vital elements in the evaluation of bug-fixing techniques since through them a researcher can conduct different kinds of tests covering various software quality aspects. These diverse and well-controlled testing environments enable the testing of bug-fixing methods in the most comprehensive manner possible, which is the result of the benchmarking process. Therefore, these methods can ultimately help to improve the software systems' functionality and fault tolerance.

\subsection{Common Gaps and Limitations in Modern Debugging Tools and Benchmarks}
Automated program repair (APR) as well as code generation are some of the powerful tools provided by modern debugging tools and benchmarks. Nevertheless, some unreliability and limitations still exist, preventing them from being more widely used and effective. Such limitations consist of questionnaire issues, necessary datasets or benchmarks, as well as difficulty dealing with complex bugs and program repair scenarios.
\begin{enumerate}
    \item
    \textbf{Generalization and Dataset Overfitting:} A key obstacle lies in the generalization of the outcomes for large and diverse populations of programs. A lot of APR systems and tools have been benchmarked based on a limited number of benchmarks, which might not fully reflect the complexity of real-world coding problems. To illustrate, tools like QFuzz utilize the fuzzing approach to discover side-channel vulnerabilities, however, challenges arise because of the randomness and incompleteness of fuzzing, so that the bug detection is not comprehensive \cite{provenfix2024}. Along with that, it is possible that the results may not be applicable to other datasets that were not evaluated, as it was in the case of tools that were tested on datasets such as Defects4J \cite{fuzzing2024}. These instruments are likely to cause overfitting when tested on these specific benchmarks. Thus, their outcomes in other contexts may be different \cite{mixtral2024}.
    
    \item
    \textbf{Bias in Tool Selection and Limited Benchmarks:} Selection bias is one of the main problems faced by contemporary debugging frameworks. In such cases the tools and benchmarks are usually the most convenient or popular ones rather than diverse sets that can cover many real-world bugs. Consequently, this can produce incorrect results and produce imperfect evaluations. A case in point is the fact that some frameworks examine only widely used benchmarks such as Defects4J and focus on specific languages like Java, thus neglecting the possible other issues \cite{evolutionary_testing}. Besides, the concern is that the dataset only consists of language-specific datasets like CodeSearchNet which would limit the results across the different programming languages \cite{deepseek2024}.

    \item
    \textbf{Non-Determinism and Overfitting in Repair Tools:} Numerous current repair systems, for example, EVO REPAIR, have non-deterministic features that can generate multiple different results in separate runs, which makes it difficult to replicate the results consistently. This results in problems such as overfitting, where tools generate "plausible" patches that pass tests but are still possibly wrong \cite{evolutionary_testing}. Overfitting is particularly problematic in the case of test case evaluation, as even test adequacy does not guarantee correctness \cite{evolutionary_testing}.

    \item
    \textbf{Limited Handling of Security and Non-Test Bugs:} APR capabilities are still developing in fixing test case failures, but the tools do not have the same ability in dealing with security vulnerabilities and non-test bugs. Many tools are concentrating on bugs that lead to test failures only, while they are ignoring the problems containing memory safety bugs or side-channel attacks, which are significant as well \cite{dear2024}. Additionally, APR tools are finding it hard to deal with the fixing of memory safety bugs like buffer overflows, and programming languages such as C and C++ are especially hard to use in this case \cite{protocol_fuzzing2024}. This problem limits the scope of their applicability in the systems where robust security measures are required.
    
    \item
    \textbf{Challenges with Input Validation and Memory Bugs:} The input validation and memory safety bugs are problems that have to be solved. They are a major obstacle for APR tools. Although APR systems can handle input sanitization issues, they often do not produce patches that cover complex validation and sanitization functions thoroughly \cite{fuzzing2024}. Additionally, the fixing of memory bugs such as use-after-free or null pointer dereference, is still a problem because of the complicated memory management that exists in languages like C and C++ \cite{protocol_fuzzing2024}. Even when the fixes are generated, they may not be able to address some deeper existing problems, such as ensuring correct memory allocation techniques.

    \item
    \textbf{Limitations in Evaluation Metrics and Test Suites: } Many benchmarks often rely on evaluation metrics that are based on limited test suites, which may not always accurately reflect the complexity and as well as the diverse nature of real-world bugs. For instance, the usage of test cases from platforms like LeetCode can cause accessibility problems and biases since testers need specific accounts and the rate limits on such external platforms can restrict their ability to test effectively \cite{evolutionary_testing}. Besides, the simulated bugs adopted in some investigations may not be the exact reflection of the actual bugs faced in the production systems. Thus, the evaluation may not be accurate and realistic \cite{evolutionary_testing}.

    \item
    \textbf{Difficulties with ML-Based Approaches:} Machine learning (ML)-based strategies for APR have different problems particularly when it comes to dealing with rare or out-of-vocabulary tokens and generating large patches. For example, fixes that need multiple new statements or that have rare function names can be difficult to generate using the current ML-based methods, which makes them ineffective in more complex cases \cite{dear2024}. Moreover, the wrong predictions of ML models can make the correction of the wrong statements such that it becomes a compound issue of the patch generation error \cite{dear2024}.
    \end{enumerate}

\section{How do different AI models and tools compare in handling various code-related tasks? What are the main strengths and weaknesses of these models?}
In recent years, various AI models and tools have emerged to assist in code-related tasks, each demonstrating unique strengths and limitations. The comparison of these models highlights their effectiveness across tasks such as code completion, bug fixing, code summarization, and code translation or refactoring. By analyzing models like Codex \cite{openai2021}, CodeT5 \cite{codet5_2021}, GraphCodeBERT \cite{graphcodebert2021}, WizardCoder \cite{wizardlm2024}, Magicoder \cite{magicoder2024}, Mixtral \cite{mixtral2024}, Smaug \cite{smaug2024}, Zephyr \cite{zephyr2024}, DeepSeek-Coder \cite{deepseek2024}, OpenCodeInterpreter \cite{opencodeinterpreter2024}, Phind \cite{phind_codellama2023}, StarCoder2 \cite{starcoder2024}, SPT Code \cite{spt_code2024}, and GPT-4, we can better understand how these tools excel in specific areas and where they fall short.\\
\\
The speed and fluency in completing codes are the main traits of Codex (OpenAI) \cite{openai2021} which has worldwide recognition, especially when it is used in GitHub Copilot. A system that gives effective suggestions in real-time, which thereby increases the productivity of developers, considerably. The drawback, however, is that Codex normally lacks accuracy in highly complex tasks or domain-dependent code, where its performance degrades. CodeT5 \cite{codet5_2021}, developed by Salesforce, is also a reliable code reading tool, as long as it is used for smaller, focused tasks. Still, the real time performance of CodeT5 is not as fast and flexible as that of Codex, so it is often not suitable for situations which require dynamic coding. On the other hand, GraphCodeBERT \cite{graphcodebert2021} from Microsoft has the ability to process more complicated code structures and it hence performs better in tasks where there is a need to understand the dependencies. However, this additional complexity comes at the cost of slower processing and the need for more resources and thus it is not suitable nor practical to use it for rapid code completion tasks.
Two of the more recent models, Phind.com/CodeLlama (34B) \cite{phind_codellama2023} and DeepSeek-Coder (34B and 33B) \cite{deepseek2024}, are models which standout. Phind's fine-tuning on Meta's Code Llama enables it to outperform GPT-4 in HumanEval benchmark tests, thus confirming the model's high accuracy in generating code from prompts. DeepSeek-Coder's "Fill-In-Middle" (FIM) training greatly enhances its ability to complete unfinished code snippets, especially across multiple languages such as Python and Java. Conversely, StarCoder2 (15B) \cite{starcoder2024} displays better performance in tasks such as HumanEval+ due to a higher mastery of code in 16 different languages as compared to CodeLlama-34B. Nonetheless, StarCoder2 has troubles when it comes to C++ code and it is not that successful in fill-in-the-middle tasks where DeepSeek-Coder has good performance.
Meanwhile, models like Smaug (72B) \cite{smaug2024} and Zephyr (7B) \cite{zephyr2024} also play an important role in code completion. Smaug is a model that has the best performance on the HuggingFace Open LLM Leaderboard with 80.48\% accuracy, leaving other models behind in terms of parameters. Zephyr, which has an AlpacaEval score of 90.6\%, is a model that is one of the best in bug detection and scalability but still lags behind GPT-4 or Claude 3.5 in complex logic handling.
\\
\\
Another area that the models perform differently is bug fixing. Codex’s \cite{openai2021} fluency makes it suitable for fixing common bugs quickly, but its ability is less effective when it comes to rare or highly contextual bugs, often missing subtle logic errors. CodeT5 \cite{codet5_2021}, on the other hand, is very exact in rectifying bugs in localized codebases due to its task-specific training, but it has trouble in real-time debugging, where Codex’s flexibility is an advantage. GraphCodeBERT \cite{graphcodebert2021} stands out in bug detection and fixing when structural dependencies are involved making it a suitable option for detecting bugs in difficult codebases. Nonetheless, similar to CodeT5, it is short in the real-time flexibility required for the fast debugging.
Phind \cite{phind_codellama2023} and DeepSeek-Coder \cite{deepseek2024} are also the two major ones in bug fixing. Phind’s augmented dataset enables it to give excellent results in the identification and solving of bugs, and even though it stumbles when it comes to specialized or domain-specific issues, it is still very strong. DeepSeek-Coder, above all, on benchmarks like Defects4J and HumanEval, shines in the detection of smaller logic errors. Regardless of that, its performance varies across different programming languages, particularly in languages such as Bash, where its poor performance compared to other models like GPT-4 is higher than others.
Magicoder (6.7B) \cite{magicoder2024} and WizardCoder (34B) \cite{wizardlm2024}, two strong models generally, exhibit some weaknesses in debugging tasks. Magic Coder is the leader in a certain benchmark over GPT-3.5 and WizardCoder, but it is not as good as CodeLlama-Python (34B) when it comes to large scale bug fixing tasks. WizardCoder is not able to catch up with GPT-4 in debugging, bug localization, and code generation, even with well-engineered prompts, so this is its biggest problem. This confirms the fact that the larger, more complicated models like GPT-4, which still lead in the handling of complex bugs due to their thorough understanding of the larger context, have the advantage over the simpler ones.
\\
\\
Regarding code summarization, CodeT5 \cite{codet5_2021} is very efficient because it produces long and coherent summaries that demonstrate a solid understanding of the code’s structure. Thus, it is suitable for structured and localized codebases. GraphCodeBERT \cite{graphcodebert2021} is really good at summarizing code with difficult dependencies which is based on graph technology that it uses to give context-aware explanations but it can be slow on large datasets. Codex is good at summary tasks, howler delivering crisp summaries for typical development environments but sometimes lacking the required level of abstraction in case of large or complex codebases.
GPT-4, because of its superior natural language skills, is very good at writing summaries of long and complicated pieces of code, usually providing more information than other models such as Codex or CodeT5. Though this has the advantage of being resource efficient, it is overall slower in performance for summarizing small or less complex codebases. Models such as SPT Code \cite{spt_code2024} are superior to other models such as GraphCodeBERT and CodeBERT in terms of tasks like code summarization, even though there are still concerns about data bias and generalizability.
StarCoder2 \cite{starcoder2024} is impressive in the areas of code summarization and machine-proof security, yet has its drawbacks in generating insecure code at high parameter sizes, such as 15B. This is a constraint when contrasted to smaller and more secure models like StableCode-3B, though StarCoder2 still has the overall better performance.
\\ \\
From the point of view of Codex\cite{openai2021} and GPT-4, both of them are the best in code translation and refactoring. Codex is especially good for basic translations of codes between different programming languages, whereas GPT-4 provides more accurate translations between different platforms, but speed is the downside. CodeT5 \cite{codet5_2021} is also strong in this area, effectively converting the code between different languages, yet preserving its functionality. However, it may be short of grasping some delicate contextual indications which a model such as GPT-4 delivers. GraphCodeBERT’s deep grasp of the structure of the code gives it a great leverage for tasks requiring the dependency and data flow knowledge, nonetheless, its slower processing speed and high resource decision make it less fit for the real-time refactoring tasks \cite{graphcodebert2021}.
Phind.com/CodeLlama \cite{phind_codellama2023} and DeepSeek-Coder \cite{deepseek2024} are also the best in code translation, particularly with multi-language support. Phind comes from its Code Llama source, it is capable of nicely handling multiple programming languages, however, it may not perform very well with languages that have extremely complex or sophisticated features. DeepSeek-Coder’s repository-level deduplication guarantees that only concise and relevant outputs are produced, which in turn increases its performance in code translation and refactoring, but its limitations still remain for niche programming languages.\\
\\
As for multilingual support, Mixtral \cite{mixtral2024} and Phind \cite{phind_codellama2023} are both the best among others for their capability of dealing with multilingual datasets in an efficient way. Notably, Mixtral beats Claude-2.1 and Llama 2 70B on instruction fine-tuning and multilingual tasks. Phind obtains high scores due to its extensive fine-tuning on programming problems for strong performance across languages. Nevertheless, Smaug-72B \cite{smaug2024} has been mainly tested on English-language datasets, which is its weakness and it cannot perform well in multilingual environments, despite its leaderboard success. GPT-4 and GraphCodeBERT \cite{graphcodebert2021} are great in general tasks but when it comes to handling multiple languages, they become more resource-intensive.
In conclusion, every model has its own strengths and weaknesses in regard to the particular coding task in question. Codex is the best at code completion and fast debugging, while GPT-4 is the king of complex tasks with unmatched precision and depth, but it has speed and resource efficiency issues. CodeT5 and GraphCodeBERT are the task-specific models, those are performing best in the structured environments but being real-time adaptable is not their strength. New models like Phind, DeepSeek-Coder, and StarCoder2 are breaking the barriers of what AI can do in terms of accuracy and language support, however, they still have difficulties in areas like resource intensity and security. By knowing the subtle differences in these models, developers, and researchers will be able to choose the most suitable tools for their particular coding needs.

\begin{figure}[htp]
    \centering
    \includegraphics[width=10cm]{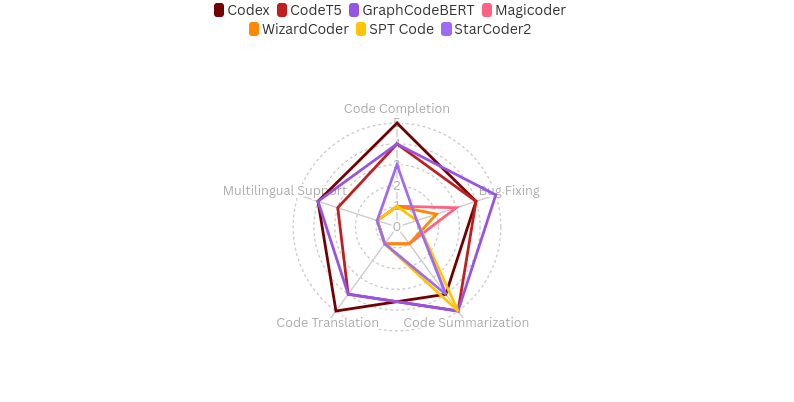}
    \caption{A diagram illustrating how different code generation models do on varying code-related tasks}
    \label{fig:graph}
\end{figure}

\begin{figure}[htp]
    \centering
    \includegraphics[width=10cm]{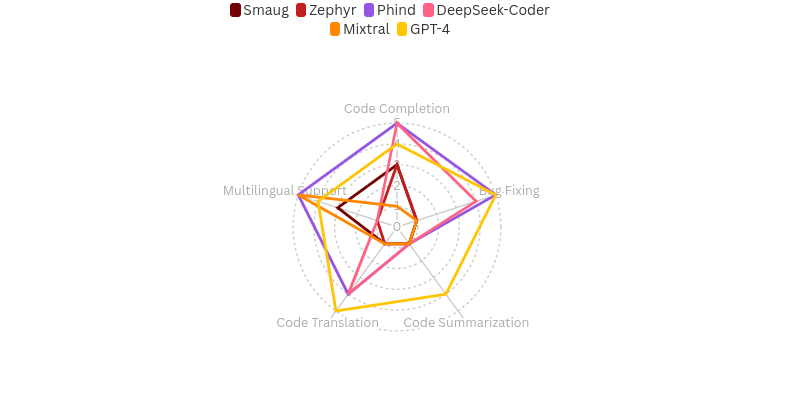}
    \caption{Some more models that were surveyed and their performance relative to others}
    \label{fig:graph}
\end{figure}

\section{How do AI models trained on general language datasets, specialized code datasets, or self-supervised learning approaches differ in their ability to handle code-related tasks?}
The AI models’ aptness to effectively tackle code-related tasks has a relationship with the way they've been trained. Different models have different approaches to the training techniques that raise the level of their performance in such areas as code generation, debugging, and semantic understanding. While many language models are pre-trained on datasets that are commonly used, which enables them to have a more general linguistic knowledge, other models are concentrated on specialized datasets to build their coding capabilities. Besides, it is a fact that some models have self-supervised or bootstrapped methods to improve their outputs, and the rest have their own distinct methodologies. This diversity comes with varying strengths and weaknesses across the models that in turn make a difference of the models in their applications in the real-world.\\ \\
Models like Codex \cite{openai2021}, CodeT5 \cite{codet5_2021}, GraphCodeBERT \cite{graphcodebert2021}, and Phind \cite{phind_codellama2023} have the commonality of being pre-trained on general language datasets. In this regard, Codex that powers GitHub Copilot and is integrated into the OpenAI API is one of the most notable names for coping with code-related tasks. It was originally fine-tuned from a collection of GPT models containing 12 billion parameters and trained on natural language data. Then, the code issue was dealt with using publicly available code from GitHub. This in-depth training enables Codex to win by a large margin over GPT-3 and GPT-J on HumanEval benchmarks because of many of their features such as Gide. One of its significant features is its iterative problem-solving capability, where repeated sampling significantly increases the likelihood of correctly solving complex tasks. The latter can be the case, however, with inconsistency in its treatment of variables when it comes to intricate code scenarios which then make its use unreliable in a code with many dependencies and complicated control flows. On the other hand, CodeT5 \cite{codet5_2021} uses an approach of a unified encoder-decoder architecture, making it possible to approach the task using both programming and natural languages, and thus it can enhance its understanding of code semantics. It makes use of developer-assigned identifiers, thus, the model improves its understanding of the code meaning which enables it to do such things as defect detection and code translation more efficiently. This bimodal learning approach makes CodeT5 to be the top performer on various tasks from the CodeXGLUE benchmark. Though it has outstanding features, it is centered on identifier recovery meaning that it would be better if it learned the structural relationships that are natural in code. GraphCodeBERT \cite{graphcodebert2021} takes a further step and it enters the world of data flow graphs so code semantics are represented not just by telling the sequence of tokens as was done by previous models. This model is pre-trained on the CodeSearchNet dataset and thus, it is capable of performing tasks like code search and clone detection due to its knowledge of the value transmission between variables. The structure-aware pre-training tasks have their own contribution, but they may be left behind in case of the more complex logic that needs the broader control flow structures. Similarly, Phind \cite{phind_codellama2023} fine-tunes the CodeLlama-34B model that is trained on a proprietary dataset of about 80,000 structured programming problems using advanced training techniques such as DeepSpeed ZeRO 3 and Flash Attention 2. The model has attained a 73.8\% pass rate on the HumanEval benchmark, thanks to a rigorous decontamination process that guarantees the integrity and validation of the dataset.\\ \\
In contrast to the ones listed above, OpenCodeInterpreter \cite{opencodeinterpreter2024}, StarCoder2 \cite{starcoder2024}, SPT Code \cite{spt_code2024}, and Magicoder \cite{magicoder2024} are models mainly pre-trained to specialized code datasets which make them more skilled at coding tasks. OpenCodeInterpreter picks up coding queries from specialized datasets, filtering them for complexity, and turning single-turn query-response pairs into multi-turn dialogues to create varied and contextual training examples. OpenCodeInterpreter's cultivation of mimicry of people was done realistically via dialogues, including but not limited to generating responses and repeating outputs until they improved their debugging skills through iterative processes. Simulated human feedback is only part of this model, which additionally benefits from purposely-produced erroneous code to make more effective debugging learning and teaching, which is particularly relevant for solving issues on platforms such as LeetCode. StarCoder2 \cite{starcoder2024} exploits a language model fine-tuned on programming data from a large scale, which allows it to convert normal text into code snippets with great success. The interactive feedback mechanism of describing the model takes the user responses and revisions into account in order to improve the output, making it more relevant and accurate. SPT Code \cite{spt_code2024} concentrates on encoding source code sequences to learn nice representations that are utilized in code completion and translation tasks, respectively. The two-step process is basically done by first pre-training on massive code datasets and then the specific tasks fine-tuning is carried out to further enhance the learning and generation capabilities. Furthermore, the last model Magicoder \cite{magicoder2024} utilizes OS code snippets through its OSS-INSTRUCT method for the creation of realistic code instructions and outperforms many state-of-the-art ones even with a smaller parameter size. Nevertheless, it is likely to inherit biases from its seed snippets which can lead to the production of low-quality outputs.\\ \\

\begin{figure}[htp]
    \centering
    \includegraphics[width=10cm]{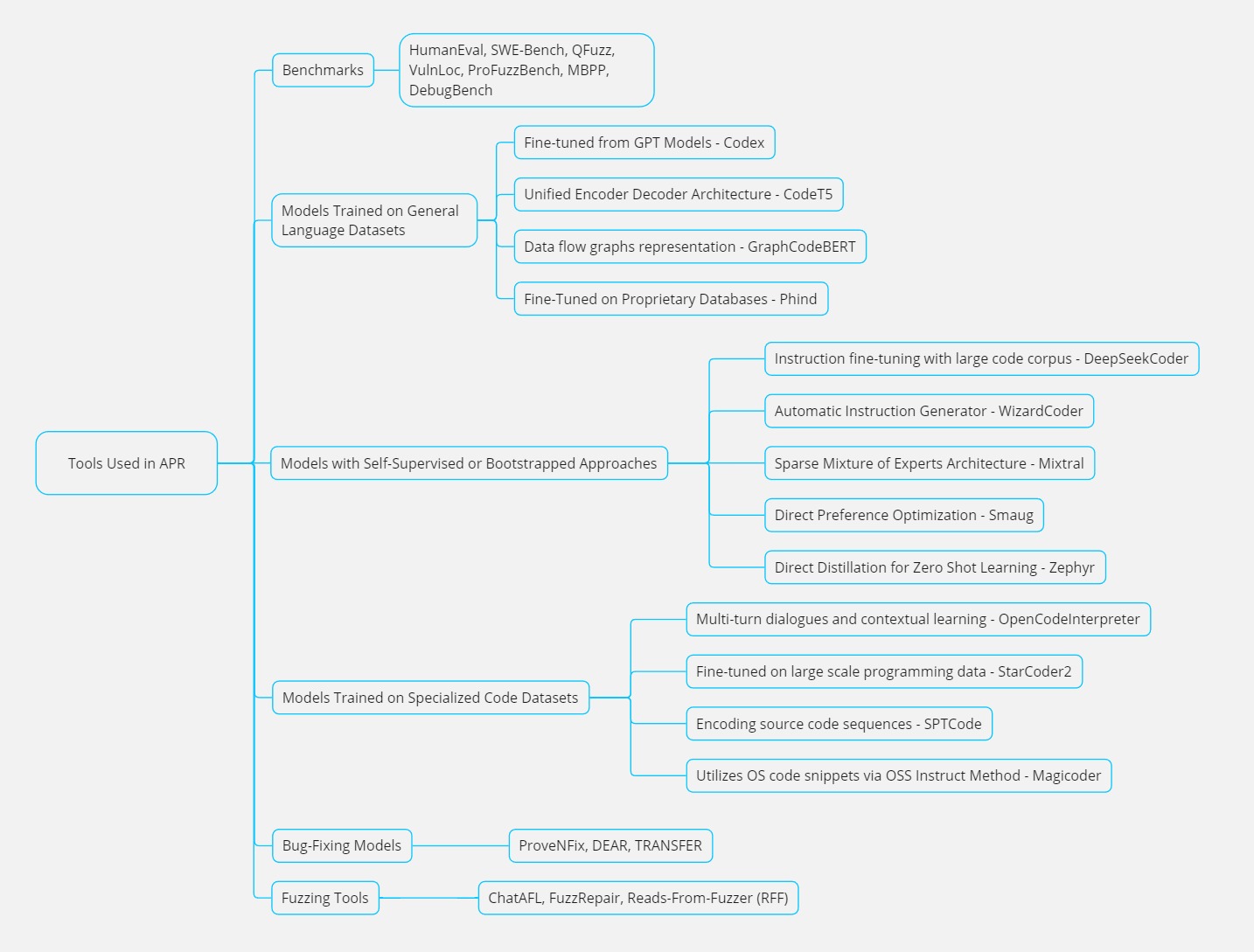}
    \caption{Learning techniques for different AI models}
    \label{fig:graph}
\end{figure}

Furthermore, models like DeepSeek-Coder\cite{deepseek2024}, WizardCoder \cite{wizardlm2024}, Mixtral \cite{mixtral2024}, and Smaug \cite{smaug2024} are equipped with self-supervised or bootstrapped methods that are augmented with innovative methodologies. DeepSeek-Coder is a specially configured fine-tuned model that is used for programming tasks, it is pre-trained on a large corpus of open-source code to increase its performance on a variety of coding tasks. Its methodology involves instruction fine-tuning with billions of tokens to respond effectively to specific coding tasks and generate meaningful outputs. The model is robust in predicting and completing code snippets with the help of Fill-In-the-Middle training approach which is a method that allows the model to learn from its outputs. With a large context window of 128K tokens, it easily tackles long coding projects. WizardCoder \cite{wizardlm2024} has introduced the Evol-Instruct method, which is the automatic generation of complex instructions. This way, the human effort is minimized and the performance on difficult tasks is increased. Despite the fact that it offers improvements over models like Alpaca and Vicuna, it may be lacking in some areas when compared to ChatGPT. Mixtral \cite{mixtral2024} uses a Sparse Mixture of Experts architecture and tuning of parameters to optimize performance by activating only some of its 47 billion parameters for inference. Thus, it acquires a unique advantage in both mathematics and code generation problems while conserving resources. Nevertheless, the intricate nature of its structure may lead to troubles concerning balancing the loads on several GPUs. Smaug improves the model's performance with the help of Direct Preference Optimization (DPO) which is a technique that refines the responses by using pairs of outputs that are preferred and dispreferred. Its DPO-Positive approach improves alignment and performance, though it may still struggle with preference datasets featuring low edit distances.\\ \\
Finally, models like Zephyr \cite{zephyr2024} propose a novel strategy of direct distillation transferring knowledge from a teacher model to a student model to improve alignment with human intentions. This methodology is put forward zero-shot learning which enables the pupil model to generalize properly and thus do well on tasks that are not taught to it explicitly. By collecting human feedback, Zephyr modifies its behavior to be more interpretable which lets it be used for various language tasks. All in all, these different methods show the ever-changing progress in AI models in programming tasks, and this is because different training techniques give different results and performance to the models.

\section{Conclusion}
In this paper, we’ve covered recent literature on the utilization of LLMs and AI in the field of automated program repair and code generation. This paper serves to summarize and curate the most up-to-date research on the subject and assist researchers in getting familiar with the state-of-the-art developments in the realm of completing code-related tasks using artificial intelligence, the use cases where this can be effective and the challenges which are currently being worked on in this field. Furthermore, we’ve also provided comparisons among all the tools that we included in this survey to assist in selecting the best open-source models to work on for making improvements in other use cases which currently don’t have reliable fixes made by AI tools.

\bibliographystyle{ACM-Reference-Format}
\bibliography{sample-base}


\begin{thebibliography}{27}


\ifx \showCODEN    \undefined \def \showCODEN     #1{\unskip}     \fi
\ifx \showDOI      \undefined \def \showDOI       #1{#1}\fi
\ifx \showISBNx    \undefined \def \showISBNx     #1{\unskip}     \fi
\ifx \showISBNxiii \undefined \def \showISBNxiii  #1{\unskip}     \fi
\ifx \showISSN     \undefined \def \showISSN      #1{\unskip}     \fi
\ifx \showLCCN     \undefined \def \showLCCN      #1{\unskip}     \fi
\ifx \shownote     \undefined \def \shownote      #1{#1}          \fi
\ifx \showarticletitle \undefined \def \showarticletitle #1{#1}   \fi
\ifx \showURL      \undefined \def \showURL       {\relax}        \fi
\providecommand\bibfield[2]{#2}
\providecommand\bibinfo[2]{#2}
\providecommand\natexlab[1]{#1}
\providecommand\showeprint[2][]{arXiv:#2}

\bibitem[phi(2023)]%
        {phind_codellama2023}
 \bibinfo{year}{2023}\natexlab{}.
\newblock \showarticletitle{Phind-CodeLlama}.
\newblock \bibinfo{journal}{\emph{Phind Technical Report}}
  (\bibinfo{year}{2023}).
\newblock
\urldef\tempurl%
\url{https://www.phind.com/blog/code-llama-beats-gpt4}
\showURL{%
\tempurl}


\bibitem[et~al.(2024a)]%
        {starcoder2024}
\bibfield{author}{\bibinfo{person}{Anton~Lozhko et al.}}
  \bibinfo{year}{2024}\natexlab{a}.
\newblock \showarticletitle{StarCoder 2 and The Stack v2: The Next Generation}.
\newblock \bibinfo{journal}{\emph{arXiv preprint}} (\bibinfo{year}{2024}).
\newblock
\urldef\tempurl%
\url{https://arxiv.org/abs/2402.19173}
\showURL{%
\tempurl}


\bibitem[et~al.(2024b)]%
        {mixtral2024}
\bibfield{author}{\bibinfo{person}{Albert Q.~Jiang et al.}}
  \bibinfo{year}{2024}\natexlab{b}.
\newblock \showarticletitle{Mixtral of Experts}.
\newblock \bibinfo{journal}{\emph{arXiv preprint}} (\bibinfo{year}{2024}).
\newblock
\urldef\tempurl%
\url{https://arxiv.org/abs/2401.04088}
\showURL{%
\tempurl}


\bibitem[et~al.(2024c)]%
        {zephyr2024}
\bibfield{author}{\bibinfo{person}{Lewis~Tunstall et al.}}
  \bibinfo{year}{2024}\natexlab{c}.
\newblock \showarticletitle{ZEPHYR: DIRECT DISTILLATION OF LM ALIGNMENT}.
\newblock \bibinfo{journal}{\emph{arXiv preprint}} (\bibinfo{year}{2024}).
\newblock
\urldef\tempurl%
\url{https://arxiv.org/pdf/2310.16944}
\showURL{%
\tempurl}


\bibitem[Gao(2024)]%
        {tabbyml2024}
\bibfield{author}{\bibinfo{person}{Lucy Gao}.} \bibinfo{year}{2024}\natexlab{}.
\newblock \bibinfo{title}{TabbyML}.
\newblock \bibinfo{howpublished}{Online}.
\newblock
\urldef\tempurl%
\url{https://github.com/TabbyML/tabby}
\showURL{%
\tempurl}


\bibitem[Guo et~al\mbox{.}(2021)]%
        {graphcodebert2021}
\bibfield{author}{\bibinfo{person}{Daya Guo}, \bibinfo{person}{Shuo Ren},
  \bibinfo{person}{Shuai Lu}, \bibinfo{person}{Long Zhou},
  \bibinfo{person}{Junjie Huang}, \bibinfo{person}{Daxin Jiang},
  \bibinfo{person}{Shuming Shi}, \bibinfo{person}{Huanbo Luan}, {and}
  \bibinfo{person}{Ming Zhou}.} \bibinfo{year}{2021}\natexlab{}.
\newblock \showarticletitle{GraphCodeBERT: Pre-training Code Representations
  with Data Flow}. In \bibinfo{booktitle}{\emph{International Conference on
  Learning Representations (ICLR)}}.
\newblock
\urldef\tempurl%
\url{https://arxiv.org/abs/2009.08366}
\showURL{%
\tempurl}


\bibitem[Guo et~al\mbox{.}(2024)]%
        {deepseek2024}
\bibfield{author}{\bibinfo{person}{Daya Guo}, \bibinfo{person}{Qihao Zhu},
  \bibinfo{person}{Dejian Yang}, \bibinfo{person}{Zhenda Xie},
  \bibinfo{person}{Kai Dong}, \bibinfo{person}{Wentao Zhang},
  \bibinfo{person}{Guanting Chen}, \bibinfo{person}{Xiao Bi},
  \bibinfo{person}{Y. Wu}, \bibinfo{person}{Y.K. Li}, \bibinfo{person}{Fuli
  Luo}, \bibinfo{person}{Yingfei Xiong}, {and} \bibinfo{person}{Wenfeng
  Liang}.} \bibinfo{year}{2024}\natexlab{}.
\newblock \showarticletitle{Deepseek-coder}.
\newblock \bibinfo{journal}{\emph{arXiv preprint}} (\bibinfo{year}{2024}).
\newblock
\urldef\tempurl%
\url{https://arxiv.org/abs/2401.14196}
\showURL{%
\tempurl}


\bibitem[Lee et~al\mbox{.}(2024)]%
        {multi_agent2024}
\bibfield{author}{\bibinfo{person}{Cheryl Lee}, \bibinfo{person}{Chunqiu~Steven
  Xia}, \bibinfo{person}{Jen tse Huang}, \bibinfo{person}{Zhouruixin Zhu},
  \bibinfo{person}{Lingming Zhang}, {and} \bibinfo{person}{Michael~R. Lyu}.}
  \bibinfo{year}{2024}\natexlab{}.
\newblock \showarticletitle{A Unified Debugging Approach via LLM-Based
  Multi-Agent Synergy}.
\newblock \bibinfo{journal}{\emph{arXiv preprint}} (\bibinfo{year}{2024}).
\newblock
\urldef\tempurl%
\url{https://arxiv.org/pdf/2404.17153}
\showURL{%
\tempurl}


\bibitem[Li et~al\mbox{.}(2024)]%
        {dear2024}
\bibfield{author}{\bibinfo{person}{Yi Li}, \bibinfo{person}{Shaohua Wang},
  {and} \bibinfo{person}{Tien~N. Nguyen}.} \bibinfo{year}{2024}\natexlab{}.
\newblock \showarticletitle{DEAR: A Novel Deep Learning-based Approach for
  Automated Program Repair}.
\newblock \bibinfo{journal}{\emph{arXiv preprint}} (\bibinfo{year}{2024}).
\newblock
\urldef\tempurl%
\url{https://dl.acm.org/doi/pdf/10.1145/3510003.3510177}
\showURL{%
\tempurl}


\bibitem[Lyu et~al\mbox{.}(2024)]%
        {automatic_programming2024}
\bibfield{author}{\bibinfo{person}{Michael~R. Lyu}, \bibinfo{person}{Baishakhi
  Ray}, \bibinfo{person}{Abhik Roychoudhury}, \bibinfo{person}{Shin~Hwei Tan},
  {and} \bibinfo{person}{Patanamon Thongtanunam}.}
  \bibinfo{year}{2024}\natexlab{}.
\newblock \showarticletitle{Automatic Programming: Large Language Models and
  Beyond}.
\newblock \bibinfo{journal}{\emph{arXiv preprint}} (\bibinfo{year}{2024}).
\newblock
\urldef\tempurl%
\url{https://arxiv.org/pdf/2405.02213}
\showURL{%
\tempurl}


\bibitem[Ma et~al\mbox{.}(2024)]%
        {software_repository2024}
\bibfield{author}{\bibinfo{person}{Yingwei Ma}, \bibinfo{person}{Qingping
  Yang}, \bibinfo{person}{Rongyu Cao}, \bibinfo{person}{Binhua Li},
  \bibinfo{person}{Fei Huang}, {and} \bibinfo{person}{Yongbin Li}.}
  \bibinfo{year}{2024}\natexlab{}.
\newblock \showarticletitle{How to Understand Whole Software Repository?}
\newblock \bibinfo{journal}{\emph{arXiv preprint}} (\bibinfo{year}{2024}).
\newblock
\urldef\tempurl%
\url{https://arxiv.org/pdf/2406.01422}
\showURL{%
\tempurl}


\bibitem[Meng et~al\mbox{.}(2024a)]%
        {protocol_fuzzing2024}
\bibfield{author}{\bibinfo{person}{Ruijie Meng}, \bibinfo{person}{Martin
  Mirchev}, \bibinfo{person}{Marcel Bohme}, {and} \bibinfo{person}{Abhik
  Roychoudhury}.} \bibinfo{year}{2024}\natexlab{a}.
\newblock \showarticletitle{Large Language Model guided Protocol Fuzzing}.
\newblock \bibinfo{journal}{\emph{arXiv preprint}} (\bibinfo{year}{2024}).
\newblock
\urldef\tempurl%
\url{https://abhikrc.com/pdf/NDSS24.pdf}
\showURL{%
\tempurl}


\bibitem[Meng et~al\mbox{.}(2024b)]%
        {fault_localization2024}
\bibfield{author}{\bibinfo{person}{Xiangxin Meng}, \bibinfo{person}{Xu Wang},
  \bibinfo{person}{Hongyu Zhang}, \bibinfo{person}{Hailong Sun}, {and}
  \bibinfo{person}{Xudong Liu}.} \bibinfo{year}{2024}\natexlab{b}.
\newblock \showarticletitle{Improving fault localization and program repair
  with deep semantic features and transferred knowledge}.
\newblock \bibinfo{journal}{\emph{arXiv preprint}} (\bibinfo{year}{2024}).
\newblock
\urldef\tempurl%
\url{https://dl.acm.org/doi/abs/10.1145/3510003.3510147}
\showURL{%
\tempurl}


\bibitem[Niu et~al\mbox{.}(2024)]%
        {spt_code2024}
\bibfield{author}{\bibinfo{person}{Changan Niu}, \bibinfo{person}{Chuanyi Li},
  \bibinfo{person}{Vincent Ng}, \bibinfo{person}{Jidong Ge},
  \bibinfo{person}{Liguo Huang}, {and} \bibinfo{person}{Bin Luo}.}
  \bibinfo{year}{2024}\natexlab{}.
\newblock \showarticletitle{SPT-Code: Sequence-to-Sequence Pre-Training for
  Learning Source Code Representations}.
\newblock \bibinfo{journal}{\emph{arXiv preprint}} (\bibinfo{year}{2024}).
\newblock
\urldef\tempurl%
\url{https://arxiv.org/abs/2201.01549}
\showURL{%
\tempurl}


\bibitem[OpenAI(2021)]%
        {openai2021}
\bibfield{author}{\bibinfo{person}{OpenAI}.} \bibinfo{year}{2021}\natexlab{}.
\newblock \showarticletitle{Evaluating Large Language Models Trained on Code}.
\newblock \bibinfo{journal}{\emph{OpenAI Technical Report}}
  (\bibinfo{year}{2021}).
\newblock
\urldef\tempurl%
\url{https://openai.com/research/evaluating-large-language-models}
\showURL{%
\tempurl}


\bibitem[Pal et~al\mbox{.}(2024)]%
        {smaug2024}
\bibfield{author}{\bibinfo{person}{Arka Pal}, \bibinfo{person}{Deep Karkhanis},
  \bibinfo{person}{Samuel Dooley}, \bibinfo{person}{Manley Roberts},
  \bibinfo{person}{Siddartha Naidu}, {and} \bibinfo{person}{Colin White}.}
  \bibinfo{year}{2024}\natexlab{}.
\newblock \showarticletitle{Smaug: Fixing Failure Modes of Preference
  Optimisation with DPO-Positive}.
\newblock \bibinfo{journal}{\emph{arXiv preprint}} (\bibinfo{year}{2024}).
\newblock
\urldef\tempurl%
\url{https://arxiv.org/pdf/2402.13228}
\showURL{%
\tempurl}


\bibitem[Ruan et~al\mbox{.}(2024a)]%
        {evolutionary_testing}
\bibfield{author}{\bibinfo{person}{Haifeng Ruan}, \bibinfo{person}{Hoang~Lam
  Nguyen}, \bibinfo{person}{Ridwan Shariffdeen}, \bibinfo{person}{Yannic
  Noller}, {and} \bibinfo{person}{Abhik Roychoudhury}.}
  \bibinfo{year}{2024}\natexlab{a}.
\newblock \showarticletitle{Evolutionary Testing for Program Repair}. In
  \bibinfo{booktitle}{\emph{International Symposium on Software Testing and
  Analysis (ISSTA)}}.
\newblock
\urldef\tempurl%
\url{https://abhikrc.com/pdf/ICST24.pdf}
\showURL{%
\tempurl}


\bibitem[Ruan et~al\mbox{.}(2024b)]%
        {timing_side_channel2024}
\bibfield{author}{\bibinfo{person}{Haifeng Ruan}, \bibinfo{person}{Yannic
  Noller}, \bibinfo{person}{Saeid Tizpaz-Niari}, \bibinfo{person}{Sudipta
  Chattopadhyay}, {and} \bibinfo{person}{Abhik Roychoudhury}.}
  \bibinfo{year}{2024}\natexlab{b}.
\newblock \showarticletitle{Timing Side-Channel Mitigation via Automated
  Program Repair}.
\newblock \bibinfo{journal}{\emph{2024 International Conference on Software
  Engineering (ICSE)}} (\bibinfo{year}{2024}).
\newblock
\urldef\tempurl%
\url{https://dl.acm.org/doi/pdf/10.1145/3678169}
\showURL{%
\tempurl}


\bibitem[Song et~al\mbox{.}(2024)]%
        {provenfix2024}
\bibfield{author}{\bibinfo{person}{Yahui Song}, \bibinfo{person}{Xiang Gao},
  \bibinfo{person}{Wenhua Li}, \bibinfo{person}{Wei-Ngan Chin}, {and}
  \bibinfo{person}{Abhik Roychoudhury}.} \bibinfo{year}{2024}\natexlab{}.
\newblock \showarticletitle{ProveNFix: Temporal Property-Guided Program
  Repair}. In \bibinfo{booktitle}{\emph{International Conference on Software
  Engineering (ICSE)}}.
\newblock
\urldef\tempurl%
\url{https://dl.acm.org/doi/pdf/10.1145/3643737}
\showURL{%
\tempurl}


\bibitem[Tian et~al\mbox{.}(2024)]%
        {evaluating_debugging2024}
\bibfield{author}{\bibinfo{person}{Runchu Tian}, \bibinfo{person}{Yining Ye},
  \bibinfo{person}{Yujia Qin}, \bibinfo{person}{Xin Cong},
  \bibinfo{person}{Yankai Lin}, \bibinfo{person}{Yinxu Pan},
  \bibinfo{person}{Yesai Wu}, \bibinfo{person}{Haotian Hui},
  \bibinfo{person}{Weichuan Liu}, \bibinfo{person}{Zhiyuan Liu}, {and}
  \bibinfo{person}{Maosong Sun}.} \bibinfo{year}{2024}\natexlab{}.
\newblock \showarticletitle{Evaluating Debugging Capability of Large Language
  Models}.
\newblock \bibinfo{journal}{\emph{arXiv preprint}} (\bibinfo{year}{2024}).
\newblock
\urldef\tempurl%
\url{https://arxiv.org/pdf/2401.04621}
\showURL{%
\tempurl}


\bibitem[Wang et~al\mbox{.}(2021)]%
        {codet5_2021}
\bibfield{author}{\bibinfo{person}{Yue Wang}, \bibinfo{person}{Weishi Wang},
  \bibinfo{person}{Shafiq Joty}, {and} \bibinfo{person}{Steven~C.H. Hoi}.}
  \bibinfo{year}{2021}\natexlab{}.
\newblock \showarticletitle{CodeT5: Identifier-aware Unified Pre-trained
  Encoder-Decoder Models for Code Understanding and Generation}. In
  \bibinfo{booktitle}{\emph{Proceedings of the 2021 Conference on Empirical
  Methods in Natural Language Processing (EMNLP)}}.
\newblock
\urldef\tempurl%
\url{https://arxiv.org/abs/2109.00859}
\showURL{%
\tempurl}


\bibitem[Wei et~al\mbox{.}(2024)]%
        {magicoder2024}
\bibfield{author}{\bibinfo{person}{Yuxiang Wei}, \bibinfo{person}{Zhe Wang},
  \bibinfo{person}{Jiawei Liu}, \bibinfo{person}{Yifeng Ding}, {and}
  \bibinfo{person}{Lingming Zhang}.} \bibinfo{year}{2024}\natexlab{}.
\newblock \showarticletitle{Magicoder: Empowering Code Generation with
  OSS-Instruct}.
\newblock \bibinfo{journal}{\emph{arXiv preprint}} (\bibinfo{year}{2024}).
\newblock
\urldef\tempurl%
\url{https://arxiv.org/abs/2312.02120}
\showURL{%
\tempurl}


\bibitem[Wolff et~al\mbox{.}(2024)]%
        {greybox_fuzzing2024}
\bibfield{author}{\bibinfo{person}{Dylan Wolff}, \bibinfo{person}{Zheng Shi},
  \bibinfo{person}{Gregory~J. Duck}, \bibinfo{person}{Umang Mathur}, {and}
  \bibinfo{person}{Abhik Roychoudhury}.} \bibinfo{year}{2024}\natexlab{}.
\newblock \showarticletitle{Greybox Fuzzing for Concurrency Testing}. In
  \bibinfo{booktitle}{\emph{Proceedings of the 2024 ACM Conference on Computer
  and Communications Security (CCS)}}.
\newblock
\urldef\tempurl%
\url{https://dl.acm.org/doi/pdf/10.1145/3620665.3640389}
\showURL{%
\tempurl}


\bibitem[Xu1 et~al\mbox{.}(2024)]%
        {wizardlm2024}
\bibfield{author}{\bibinfo{person}{Can Xu1}, \bibinfo{person}{Qingfeng Sun},
  \bibinfo{person}{Kai Zheng}, \bibinfo{person}{Xiubo Geng},
  \bibinfo{person}{Pu Zhao}, \bibinfo{person}{Jiazhan Feng},
  \bibinfo{person}{Chongyang Tao}, \bibinfo{person}{Qingwei Lin}, {and}
  \bibinfo{person}{Daxin Jiang}.} \bibinfo{year}{2024}\natexlab{}.
\newblock \showarticletitle{WizardLM: Empowering Large Language Models to
  Follow Complex Instructions}.
\newblock \bibinfo{journal}{\emph{arXiv preprint}} (\bibinfo{year}{2024}).
\newblock
\urldef\tempurl%
\url{https://arxiv.org/abs/2304.12244}
\showURL{%
\tempurl}


\bibitem[Zhang et~al\mbox{.}(2024)]%
        {fuzzing2024}
\bibfield{author}{\bibinfo{person}{Yuntong Zhang}, \bibinfo{person}{Ridwan
  Shariffdeen}, \bibinfo{person}{Gregory~J. Duck}, \bibinfo{person}{Jiaqi Tan},
  {and} \bibinfo{person}{Abhik Roychoudhury}.} \bibinfo{year}{2024}\natexlab{}.
\newblock \showarticletitle{Program Repair by Fuzzing over Patch and Input
  Space}. In \bibinfo{booktitle}{\emph{International Symposium on Software
  Testing and Analysis (ISSTA)}}.
\newblock
\urldef\tempurl%
\url{https://arxiv.org/pdf/2308.00666}
\showURL{%
\tempurl}


\bibitem[Zheng et~al\mbox{.}(2024)]%
        {opencodeinterpreter2024}
\bibfield{author}{\bibinfo{person}{Tianyu Zheng}, \bibinfo{person}{Ge Zhang},
  \bibinfo{person}{Tianhao Shen}, \bibinfo{person}{Xueling Liu},
  \bibinfo{person}{Bill~Yuchen Lin}, \bibinfo{person}{Jie Fu},
  \bibinfo{person}{Wenhu Chen}, {and} \bibinfo{person}{Xiang Yue}.}
  \bibinfo{year}{2024}\natexlab{}.
\newblock \showarticletitle{OpenCodeInterpreter: Integrating Code Generation
  with Execution and Refinement}.
\newblock \bibinfo{journal}{\emph{arXiv preprint}} (\bibinfo{year}{2024}).
\newblock
\urldef\tempurl%
\url{https://arxiv.org/abs/2402.14658}
\showURL{%
\tempurl}


\bibitem[Zhong et~al\mbox{.}(2024)]%
        {debug_like_human2024}
\bibfield{author}{\bibinfo{person}{Li Zhong}, \bibinfo{person}{Zilong Wang},
  {and} \bibinfo{person}{Jingbo Shang}.} \bibinfo{year}{2024}\natexlab{}.
\newblock \showarticletitle{Debug like a Human}. In
  \bibinfo{booktitle}{\emph{Proceedings of the 2024 ACM Conference on Computer
  and Communications Security (CCS)}}.
\newblock
\urldef\tempurl%
\url{https://arxiv.org/pdf/2402.16906}
\showURL{%
\tempurl}


\end{thebibliography}

\end{document}